\documentclass[letterpaper,11pt]{IEEEtran}
\usepackage{graphicx,times,amsmath,amssymb,subfigure,cite,url,hyperref}
\usepackage[flushleft]{threeparttable}
\usepackage[lined,ruled,commentsnumbered]{algorithm2e}
 \allowdisplaybreaks
\renewcommand{\baselinestretch}{1}

\begin{document}

\title{OMG - Emotion Challenge Solution}
\author{Yuqi Cui, Xiao Zhang, Yang Wang, Chenfeng Guo and Dongrui Wu\\
Huazhong University of Science and Technology, Wuhan, China\\
 Email: drwu@hust.edu.cn}

\maketitle

\begin{abstract}
This short paper describes our solution to the 2018 IEEE World Congress on Computational Intelligence \emph{One-Minute Gradual-Emotional Behavior Challenge}, whose goal was to estimate continuous arousal and valence values from short videos. We designed four base regression models using visual and audio features,  and then used a spectral approach to fuse them to obtain improved performance.
\end{abstract}

\begin{IEEEkeywords}
Affective computing, emotion estimation
\end{IEEEkeywords}

\section{Problem Statement}

The One-Minute Gradual-Emotional Behavior Challenge\footnote{https://www2.informatik.uni-hamburg.de/wtm/OMG-EmotionChallenge/} was a competition organized at the 2018 IEEE World Congress on Computational Intelligence\footnote{http://www.ecomp.poli.br/~wcci2018/competitions/} (IEEE WCCI 2018). The dataset was composed of 420 relatively long emotion videos with an average length of 1 minute, collected from a variety of Youtube channels. Videos were separated into clips based on utterances, and each utterance's valence and arousal levels were annotated by at least five independent subjects using the Amazon Mechanical Turk tool. The goal was to estimate the valence and arousal levels for each utterance, from modalities such as visual, audio, and text. The training dataset consisted of 2,442 utterances, validation dataset 621 utterances, and testing dataset 2,229 utterances.

The performance measure was the Congruence Correlation Coeficient (CCC). Let $N$ be the number of testing samples, $\{y_i\}_{i=1}^N$ be the true valence (arousal) levels, and $\{\hat{y}_i\}_{i=1}^N$ be the estimated valence (arousal) levels. Let $m$ and $\sigma$ be the mean and standard deviation of $\{y_i\}$, respectively, $\hat{m}$ and $\hat{\sigma}$ be the mean and standard deviation of $\{\hat{y}\}$, respectively, and $\gamma$ be the Pearson correlation coefficient between $\{y_i\}$ and $\{\hat{y}\}$. Then, the CCC is computed as:
\begin{align}
ccc=\frac{2\gamma\sigma\hat{\sigma}}{\sigma^2+\hat{\sigma}^2+(m-\hat{m})^2}
\end{align}
Clearly, $ccc\in[-1,1]$.

More information about the dataset and some baseline results can be found in \cite{Barros2016}.

\section{Our Solution and Results}

We developed four base regression models, and then aggregated their outputs by spectral meta-learner for regression (SMLR) \cite{drwuSMLR2016}.

\subsection{The CNN-Face Model}

We used the \emph{face$\_$recognition} package\footnote{https://github.com/ageitgey/face$\_$recognition} to crop out the face of the actor in each frame of an utterance, and then performed emotion analysis on the faces only. Each face image was rescaled to $80\times80\times3$ (height$\times$width$\times$channel). We extracted face features by Xception \cite{Chollet2016} with weights pre-trained on ImageNet. Each utterance gave $n$ 2048-d feature vectors, where $n$ is the number of frames. We then took the average of these $n$ 2048-d feature vectors to obtain a single 2048-d feature vector for each utterance. These features were next passed through a three-layer multi-layer perception (MLP) for regression. The hidden layer had 1024 nodes with ReLU activation, and the output layer had only one node with sigmoid activation for arousal, and linear activation for valence. Optimization of the MLP was done using Adamdelta, with dropout rate $0.25$. The validation ccc was used to determine when the training should stop.

\subsection{The CNN-Visual Model}

This model was almost identical to \emph{CNN-Face}, except that the entire frame instead of only the face was used to extract the features.

\subsection{The LSTM-Visual Model}

This regression model was inspired by the video classification model in \cite{Ng2015}. For each utterance, we down-sampled 20 frames uniformly in time (if an utterance had less than 20 frames, then the first frame was repeated to make up 20 frames), and then used InceptionV3 \cite{Szegedy2016}, pre-trained on ImageNet, to obtain a $20\times2048$ feature matrix. Next we applied multi-layer long short-term memory (LSTM) to extract the time domain information, and an MLP with 512 hidden nodes and one output node for regression. Dropout and ReLU activation were used in both LSTM and MLP.

\subsection{The SVR-Audio Model}

We first converted the .mp4 audio format to .wav format, partitioned each utterance into frames, and then extracted the following features using moving windows (window length 200, sliding distance 80):

\begin{enumerate}
\item \textbf{Low-level features}, which describe the basic properties of audio in time- and frequency- domains, including the spectral centroid, band energy radio, delta spectrum magnitude, zero crossing rate, short-time average energy, and pitch. More details about these low-level features can be found in \cite{Li2001}.
\item \textbf{Silence ratio}, which is the ratio of the amount of silence frames to the time window \cite{Chen2006}. A frame is considered as a silence frame when its root mean square is less than 50\% of the mean root mean square of the fixed-length audio fragments.
\item \textbf{MFCCs and LPCCs}. In order to combine the static and dynamic characteristics of audio signals, 12 Mel Frequency Cepstral Coefficients (MFCCs), 11 Linear Predictive Cepstral Coefficients (LPCCs), and 12 first-order differential MFCC coefficients were calculated.
\item \textbf{Formant}, which reflects the resonant frequencies of the vocal tract. Formant frequencies F1-F5 in each frame were extracted.
\end{enumerate}
We then computed the mean and/or variance of these frame-level features, resulting in a total of 76 audio features, as shown in Table~\ref{tab:audio}. These 76 features have been used in our previous research \cite{drwuACII2018}.

\begin{table}[htbp]\centering \setlength{\tabcolsep}{2mm}
\newcommand{\tabincell}[2]{\begin{tabular}{@{}#1@{}}#2\end{tabular}}
\caption{The 76 audio features.} \label{tab:audio}
\begin{tabular}{|c|c|c|} \hline
\textbf{Feature category}&\textbf{Number}&\textbf{Value} \\ \hline
\tabincell{c}{Spectral centroid,\\Band energy radio,\\Delta spectrum magnitude,\\Zero crossing rate, \\Pitch,\\Short-time average energy }& 12 & Mean, variance \\  \hline
Silence ratio &  1  &  Mean \\ \hline
\tabincell{c}{MFCC coefficients,\\Delta MFCC,\\LPCC}&\tabincell{c}{24\\12\\22}&\tabincell{c}{Mean, variance\\Mean\\Mean, variance} \\ \hline
Formant &5&	Mean \\  \hline
\end{tabular}
\end{table}

In this solution, instead of using these 76 features directly, we first clipped each feature into its $[2,98]$ percentile interval (e.g., all values smaller than 2 percentile were replaced by the value at 2 percentile, and all values larger than 98 percentile were replaced by the value at 98 percentile), normalized to $[0,1]$, and then used RReliefF \cite{Robnik2003} to sort the features according to their importance. Next, we used support vector regression (SVR) and the validation dataset to determine the appropriate number of features to use. We performed feature clipping because many features had extreme values, which significantly deteriorated the estimation performance.

\subsection{Model Fusion by SMLR}

The above base regression models were then fused by our recently developed SMLR approach\footnote{We did not use the clustering step in \cite{drwuSMLR2016} because we only had four base regression models here.} \cite{drwuSMLR2016}. SMLR first uses a spectral approach to estimate the accuracies of the base regression models on the testing dataset, and then uses a weighted average to combine the base regression models (the weights are the accuracies of the base models) to obtain the final estimates on the testing dataset.

\subsection{Results}

The validation results on the CCC and mean squared error (MSE) are shown in Table~\ref{tab:results}. Note that CNN-Visual was not used in SMLR fusion for Arousal since its performance was too low. We can observe from Table~\ref{tab:results} that:
\begin{enumerate}
\item SVR-Audio achieved better CCCs than the other three base regression models on the visual or face.
\item SMLR achieved the best performance on both CCC and MSE, suggesting the fusion was effective.
\end{enumerate}

\renewcommand{\baselinestretch}{1.2}
\begin{table}[htbp]\centering \setlength{\tabcolsep}{3mm}
\newcommand{\tabincell}[2]{\begin{tabular}{@{}#1@{}}#2\end{tabular}}
\caption{The validation results.} \label{tab:results}
\begin{tabular}{|c|c|c|c|c|} \hline
&\textbf{CCC }&\textbf{CCC } & \textbf{MSE } &\textbf{MSE } \\
\textbf{Model}&\textbf{Arousal}&\textbf{Valence} & \textbf{Arousal} &\textbf{Valence} \\ \hline
CNN-Face & 0.3214 & 0.3606 & 0.0551 & 0.1163 \\
CNN-Visual & 0.2448 & 0.3568 & 0.0515 & 0.1045 \\
LSTM-Visual & 0.3383 & 0.3694 & 0.0431 & 0.1382 \\
SVR-Audio & 0.3693 & 0.4150 & 0.0543 & 0.1089\\ \hline
SMLR & \textbf{0.3969} & \textbf{0.4411} & \textbf{0.0404} & \textbf{0.0910}\\ \hline
\end{tabular}
\end{table}
\renewcommand{\baselinestretch}{1}



\end{document}